\documentclass[letterpaper]{article} 
\usepackage{aaai23}  
\usepackage{times}  
\usepackage{helvet}  
\usepackage{courier}  
\usepackage[hyphens]{url}  
\usepackage{graphicx} 
\usepackage{natbib}  
\usepackage{caption} 
\usepackage{algorithm}
\usepackage{algorithm}
\usepackage{algorithmic}
\usepackage{subfiles}
\usepackage{amsmath}
\usepackage[dvipsnames]{xcolor}         
\usepackage{subcaption}
\usepackage{url}
\usepackage{amsfonts}
\usepackage{nicefrac}
\usepackage{booktabs}
\usepackage{blindtext}
\usepackage{multicol}
\usepackage{newfloat}
\usepackage{listings}
\DeclareCaptionStyle{ruled}{labelfont=normalfont,labelsep=colon,strut=off} 
\floatstyle{ruled}
\newfloat{listing}{tb}{lst}{}
\floatname{listing}{Listing}
\title{Deep Reinforcement Learning with Vector Quantized Encoding}
\author{
    Liang Zhang,
    Justin Lieffers,
    Adarsh Pyarelal
}
\affiliations{
    The University of Arizona \\

    Tucson, AZ 85721 USA\\
    \{liangzh, lieffers, adarsh\}@arizona.edu
}

\begin{document}

\maketitle
\begin{abstract}

Human decision-making often involves combining similar states into categories
and reasoning at the level of the categories rather than the actual states.
Guided by this intuition, we propose a novel method for clustering state
features in deep reinforcement learning (RL) methods to improve their
interpretability. Specifically, we propose a plug-and-play framework termed
\emph{vector quantized reinforcement learning} (VQ-RL) that extends classic RL
pipelines with an auxiliary classification task based on vector quantized (VQ)
encoding and aligns with policy training. The VQ encoding method categorizes
features with similar semantics into clusters and results in tighter clusters
with better separation compared to classic deep RL methods, thus enabling
neural models to learn similarities and differences between states better.
Furthermore, we introduce two regularization methods to help increase the
separation between clusters and avoid the risks associated with VQ training. In
simulations, we demonstrate that VQ-RL improves interpretability and
investigate its impact on robustness and generalization of deep RL.

    \end{abstract}

    \section{Introduction}
    
\label{sec:introduction}

\begin{figure}
        \centering
        \includegraphics[width=\columnwidth]{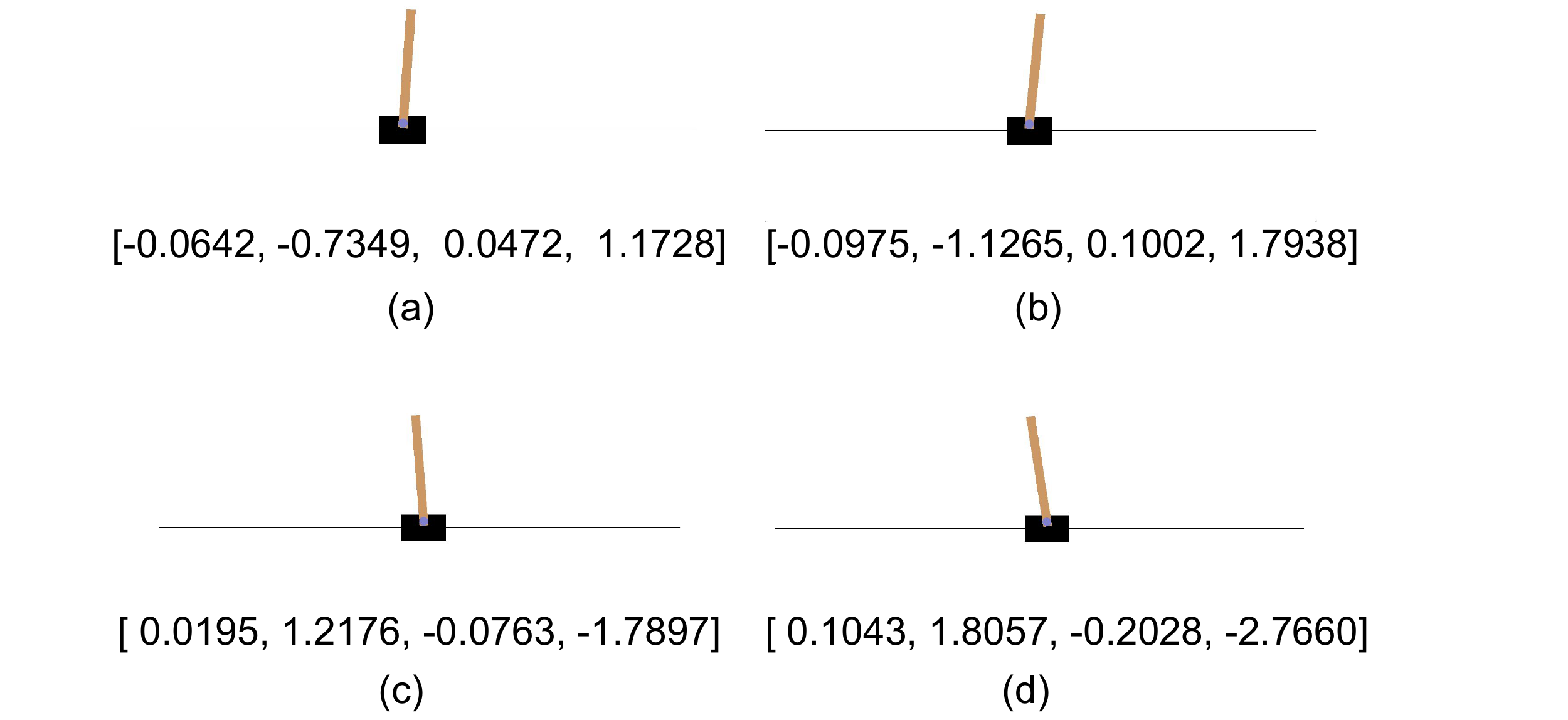}
        \caption{%
            In the CartPole game, players control a cart and move it left or
            right to keep the upper pole balanced. The vector below each figure
            represents the current state,  being: cart position (CP), cart velocity (CV), pole
            angle (PA), and pole angular velocity (PAV). Each feature is a continuous
            decimal value. In (a) and (b),  the pole deviates to the right, and
            the player needs to take the right movement action to maintain
            balance, while in (c) and (d), the cart requires movement to the
            left.
        }
        \label{fig:cartpole_demo}
    \end{figure}

  Human thinking reflects clustering
    characteristics in many modalities such as reasoning and
    planning. In the CartPole game \citep{barto1983neuronlike}, as shown in
    Figure \ref{fig:cartpole_demo}, humans cannot estimate the values of the
    components of the game state with the same precision as a
    computer - however, they can make general classifications on states and
    suggest corresponding actions. For example, when the pole tilts to the left, we need to move the cart to
    the left to keep the pole balanced, and vice versa. In this
    process, our brain does not classify the various states of the pole to the
    left with arbitrary precision, but processes similar states as one
    category. This leads one to ask if we could exploit the clustering
    properties of feature distributions in neural networks in deep reinforcement learning (RL) problems
    to (i) explore the relationships between states and (ii) infer
    relationships between clusters and actions/ policies. Doing so will help us
    understand more about the decision-making process in deep RL and the
    clustering relationships between states.

    \begin{figure}
        \centering
        \includegraphics[width=\columnwidth]{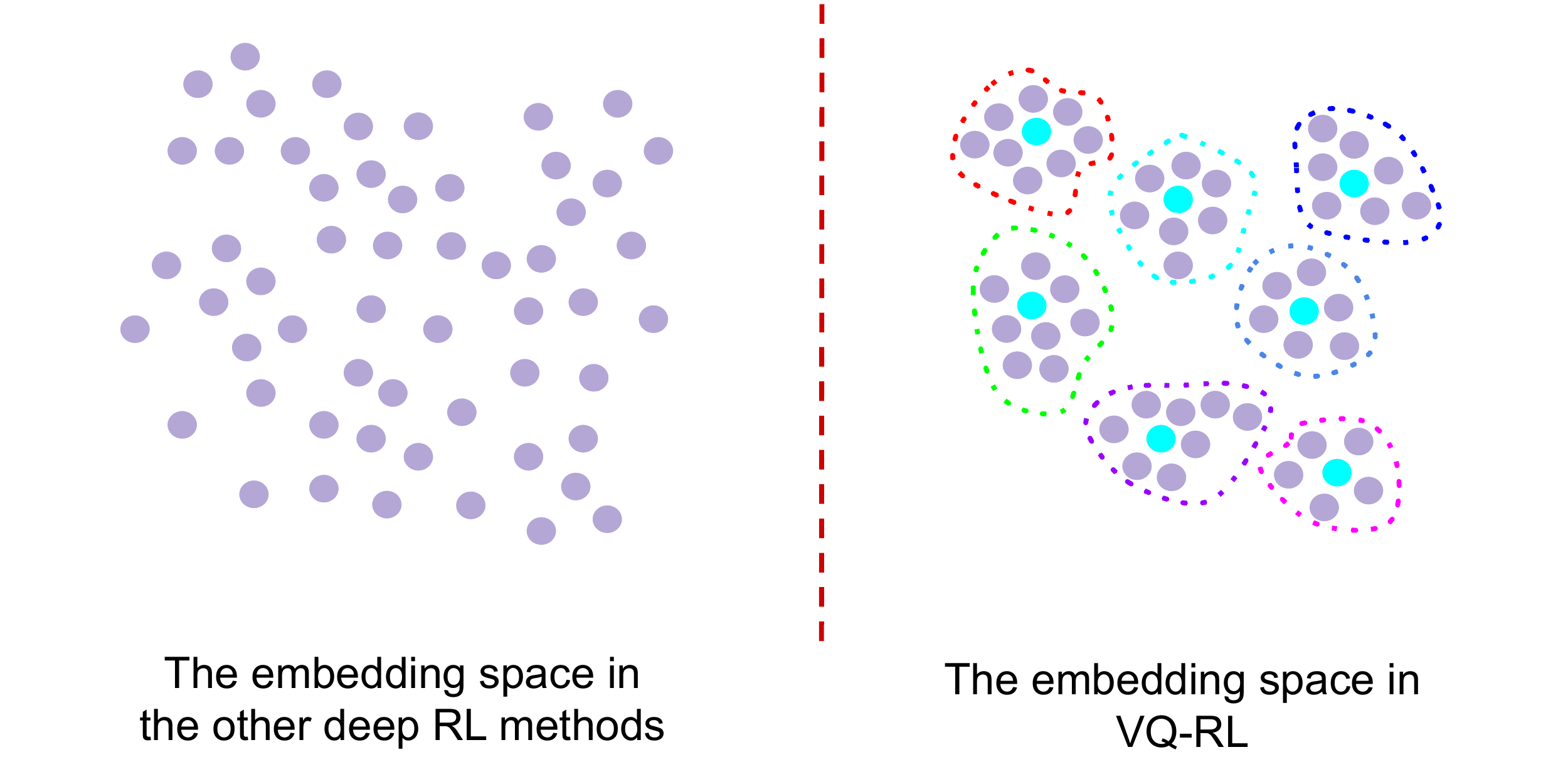}
        \caption{%
            The features (purple circles) extracted by neural networks are
            scattered in a high-dimensional continuous embedding space.
            On the other hand, the features extracted in
            VQ-RL are clustered around the codebook's candidate embeddings (light
            blue circles), and the codebook's closest embedding is used as a
            replacement embedding. Besides, the clustering is guided by policy training, i.e., states in the same cluster map the same highest probability of action in policy.
        }
        \label{fig:embedding_space}
    \end{figure}

    In the vein of understanding latent spaces of neural networks better,
    vector quantized variational autoencoders (VQ-VAE) \cite{oord2017neural}
    are a successful family of generative models that combine the variational
    autoencoder (VAE) framework with discrete latent representations through a
    novel parameterization of the posterior distribution given observations.
    VQ-VAEs provide an
    effective method for giving more access to the latent space of feature encodings. So far,
    most of the applications of VQ-VAEs have focused on generative models, such
    as generating images, audio, video, etc. \citep{oord2017neural,
    garbacea2019low, razavi2019generating, tjandra2020transformer}.
    Nevertheless, this encoding and training process can also bring us two
    advantages: First, VQ encoding generates
    latent feature spaces with improved clustering properties (see 
    Figure \ref{fig:embedding_space}), i.e., we can more effectively learn the
    commonality between features within a cluster, and better distinguish
    features between different clusters. Second, according to the t-distributed
    Stochastic Neighbor Embedding (t-SNE) \citep{van2008visualizing}
    visualizations in \citet{mnih2015human, zahavy2016graying}, in the
    embedding space with dimensionality reduction guided by RL policy training,
    state features with the same or similar semantics are more likely to be clustered together. These properties provide the potential to implement 
    our proposed clustering intuition in this study.

    In order to apply this idea to RL decision-making processes, we
    introduce a novel
    learning pipeline, named \emph{vector-quantized reinforcement learning} or VQ-RL in this work.
    In our experiments we will study the effectiveness of VQ-RL, analyzing if the improved clustering will lead to increased interpretability, robustness, and generalization.
    In summary, we make the following contributions in this paper:

\begin{itemize}
    \item{Inspired by the clustering nature of human reasoning, we propose a
        novel multi-task learning framework based on VQ encoding that improves the clustering properties and interpretability of features.}
    \item{We introduce two regularization methods to further improve the
        separation between clusters in the encoding space and avoid the risk in the codebook training that may be caused by large values.}
    \item{We design and perform three sets of experiments on a variety of domains with
        both continuous and discrete state spaces to demonstrate the effectiveness of our proposed architecture.}
\end{itemize}

    \section{Related Work}
    
Although deep neural networks have succeeded in many RL problems \cite{levine2016end, lee2019stochastic, jaderberg2019human, silver2016mastering},
    their black-box characteristics make them difficult to interpret. Besides, it is still challenging to reach human levels of robustness and
    generalization. Therefore, research on extracting
    more interpretable, robust and generalized feature representations is actively underway.

    \paragraph{Auxiliary Tasks}  Auxiliary tasks are often incorporated into
    deep RL to improve feature representation and
    performance. The unsupervised Pixel Control task,
    which predicts screen changes in discrete control environments, was applied
    in the UNREAL agent \cite{jaderberg2016reinforcement}.
    \citet{oord2018representation} proposed Contrastive Predictive Coding (CPC)
    to extract useful representations from high-dimensional data, which
    predicted the future latent space by using powerful autoregressive models.
    CPC demonstrated the ablility to learn valuable representations while also
    achieving strong performance on 3D RL problems. Predictions of Bootstrapped
    Latents (PBL) \cite{guo2020bootstrap} builds on multistep predictive
    representations of future observations and focuses on capturing structured
    information about environment dynamics, which improves the performance of
    state-of-the-art RL methods. To overcome the limitations of reward-driven feature learning in deep RL
    from images, the Augmented Temporal Contrast (ATC)
    \cite{stooke2021decoupling} method was proposed, which matches or
    outperforms end-to-end RL in most RL testbeds.  \citet{diuk2008object}
    introduced a representation based on objects and their interactions, which
    provides a natural way of modeling environments and offers significant
    generalization opportunities.  

\paragraph{Interpretability} The interpretability of
deep RL has been extensively studied
in recent years.  Programmatically Interpretable Reinforcement Learning (PIRL)
\cite{verma2018programmatically} represents policies using a high-level,
domain-specific programming language designed to generate interpretable and
verifiable agent policies. To improve the interpretability of the subtasks in
hierarchical decision-making, \cite{lyu2019sdrl} introduced symbolic planning
into RL and proposed a framework of Symbolic Deep Reinforcement Learning (SDRL)
that can handle both high-dimensional sensory inputs and symbolic planning. A soft attention model for reinforcement learning domains was introduced in
\cite{mott2019towards} to show that the model learns to query separately about
space and content (``where" vs. ``what").  Some visualization techniques
\cite{such2018atari, wang2018dqnviz, zahavy2016graying}
for feature space have been proposed for better interpretability of RL methods, which showed that features with similar semantics locate closely.
While these methods revealed that features in the embedding space have specific
clustering properties, they do not explore how these clustering properties can
be used to further enhance the performance of RL methods. VQ encoding directly acts on constructing the clustered embedding space, with VQ embeddings as the centers to attract the surrounding features into a smaller space, providing us the potential to cluster features with similar semantics.

\paragraph{Robustness and generalization} An approach for robustness to action
uncertainty was proposed in \cite{tessler2019action}, which provided a
robust policy learning method and improved performance in the
absence of perturbations.  To increase robustness to noise and adversaries,
\citet{lutjens2020certified} introduced an online certified defense to the Deep
Q-Network policy training.  \citet{wang2020reinforcement} studied deep RL
models in noisy reward scenarios and developed a robust framework for
these scenarios. Furthermore, \cite{cobbe2019quantifying} introduced CoinRun, a benchmark for
generalization in RL, and studied some factors in neural network training that
may affect generalization.  \cite{packer2018assessing} proposed a benchmark and
experimental protocol and conducted a systematic empirical study. Some specific
neural network training techniques have also been proposed
 in the literature to improve generalization
in RL \cite{lee2019network, zhang2018dissection, igl2019generalization}.

\paragraph{VQ-VAE}  In recent years, VQ-VAE has been successfully applied to
various tasks, such as high-resolution image generation
\cite{razavi2019generating}, video generation
\cite{DBLP:journals/corr/abs-2104-10157}, speech coding
\cite{DBLP:conf/icassp/GarbaceaOLLLVW19}, etc. VQ-VAE has also been introduced
into model-based deep RL problems to train transition models in some recent
works \cite{robine2020smaller}. Unlike the above papers
where VQ-VAE is utilized for different reconstruction or generation tasks, this
paper leverages VQ encoding to cluster features in RL problems and
train a policy-guided classification task based on encoded features.

    \section{Method}
    
In this section, we explain our methodology and architecture, which
builds on the VQ-VAE work proposed by \citet{oord2017neural}. Here we provide a brief
overview of the relevant VQ-VAE equations and concepts.

The VQ-VAE model starts with an encoder network that maps an input $x$ to a latent continuous representation $E(x)$. After that, $E(x)$ is quantized via the nearest neighbor look-up in the codebook, $\mathbf{e} \in R^{K\times D}$, where $K$ is the number of vectors in the codebook and $D$ is the dimension of the continuous embedding. The $K$ embedding vectors are represented as an index of the codebook, $\mathbf{e}_{i} \in R^{D}, i \in 1 \ldots K$. This nearest neighbor vector is then fed into the decoder network to reconstruct the input $x$. The following
equation summarizes the process of the quantization of VQ encoding:

\begin{equation}
    \label{vq-quantization}
    \begin{split}
        \operatorname{Quantize}(E(\mathbf{x}))=\mathbf{e}_{k},\\
        k=\underset{j}{\arg \min }\left\|E(\mathbf{x})-\mathbf{e}_{j}\right\|
    \end{split}
\end{equation}

The objective of the VQ-VAE training follows:
\begin{equation}
    \label{eq:vq-vae-loss}
    \begin{split}
        \mathcal{L}(\mathbf{x}, D(\mathbf{e}))=\|\mathbf{x}-D(\mathbf{e})\|_{2}^{2}+\|s g[E(\mathbf{x})]-\mathbf{e}\|_{2}^{2}\\
        +\beta\|s g[\mathbf{e}]-E(\mathbf{x})\|_{2}^{2}
    \end{split}
\end{equation}

\noindent
where $E$ is the encoder and $D$ is the decoder, $s g$ is a stop-gradient operator, and $\beta$ is the weight preventing the encoder outputs from fluctuating between different code vectors.

\begin{figure}
    \centering
    \includegraphics[width=\columnwidth]{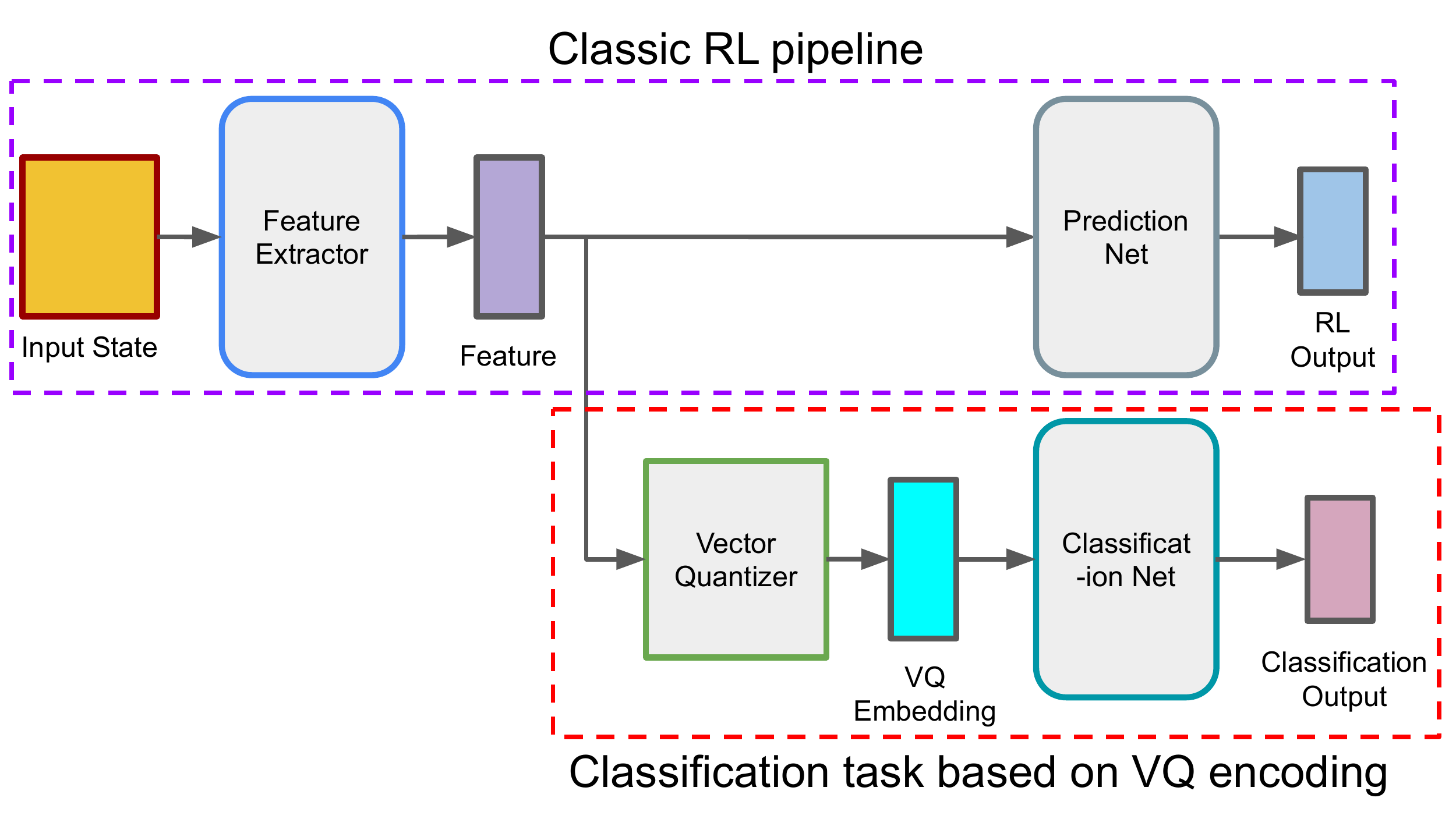}
    \caption{%
        Overview of the VQ-RL architecture. VQ-RL adds an
        auxiliary classification task based on VQ encoding to the classic
        RL pipeline. The vector quantizer re-encodes features extracted
        from the feature extractor, returning the closest embedding from
        the codebook as the output. Based on the VQ embedding, the
        classification net outputs the action with the highest probability
        from the current policy.
    }
    \label{fig:architecture}
\end{figure}


\subsection{VQ-RL}

Deep RL architectures usually contain two modules: a feature extractor and
a prediction network, whose designs
vary according to the specific
requirements of the problems being studied 
\cite{bellemare2017distributional, kaiser2019model, arulkumaran2017deep}.
To improve the extracted features' clustering properties in the embedding
space, we introduce an auxiliary task with
discretized VQ encoding to the classic RL pipeline. This auxiliary task aims to
predict the maximum possible output action under the current policy by
performing a classification task and will lead to clustering together features
with the same maximum possible output, such as the example in Figure \ref{fig:cartpole_demo}. The architecture of our model is shown in Figure \ref{fig:architecture}, and the visualization of vector quantizer is shown in Appendix.

The training process in VQ-VAE causes two effects: 1) The
encoded features learn to be as close as possible to the closest
embedding vector in the codebook, 2) the embedding vectors in the
codebook learn to be as close as possible to all the features in
their corresponding cluster. Theoretically, VQ encoding can help
achieve clustering properties without performing additional
unsupervised tasks. 
Similar to the encoding process in VQ-VAE, the vector quantizer in VQ-RL returns (i): the embedding in the codebook as the output that is closest to the input feature and
(ii): the encoding index identifying the cluster.

Nevertheless, in the reconstruction task of VQ-VAE, features are segmented
into multiple parts and VQ encoding is processed for each segment
separately. For example, in the image reconstruction task in VQ-VAE
\cite{oord2017neural}, a $32 \times 32$ embedding matrix with $K=512$ for
ImageNet is required. In order to cluster features in VQ-RL, we set the
embedding vector dimension in the codebook to be the same as the input feature size.

The most straightforward choice for the number of embeddings in the
codebook is the number of possible actions, e.g., the original
CartPole domain \cite{barto1983neuronlike} contains only two performable actions -
thus, for this domain, we would simply set the number of embeddings in the
codebook to two. However, in practice, the human brain may further divide the state
space into more regions than there are possible primitive actions
in the domain. For an instance, in the CartPole domain, although it is
necessary to perform the `move left' action both when the pole is
slightly to the left and very far to the left, we tend to
distinguish them into two types of states since it affects future
actions.  Similarly, in CoinRun, approaching a moving enemy and a
fixed obstacle requires performing the `jump' action, but we
usually treat these two cases differently. Therefore, we set the
number of embeddings to be larger than the number of performable
actions in the experiments to further explore the embedding space's
internal status in greater detail and to better emulate human
reasoning. 





\subsection{Regularization and loss function}

To improve the robustness of the codebook learning, we introduce and modify
two regularization forms from \citet{franccois2019combined} into our training. The first,
$\mathcal{L}_{d1}$ is defined as follows:

\vspace{-1em}
\begin{equation}
    \mathcal{L}_{d 1}=\sum_{i \ne j}\exp \left(-C_{d}\left\|\mathbf{e}_{i}-\mathbf{e}_{j}\right\|_{2}\right)
\end{equation}

\noindent
where $C_{d}$ is a constant. In contrast to the form in \citet{franccois2019combined}
    which considers random pairs, we calculate the sum
    loss of all possible pairs in the codebook. This prevents any two
    candidate embedding vectors from getting too close in the latent space
    and ensures the difference among codebook embedding vectors.

Similarly to the prior used in the original VAE \cite{kingma2013auto}, we penalize the values of
embedding vectors out of an $L_{\infty}$ ball of radius 1. This prevents
the risk that may be caused by large values of embedding vectors and
is expressed as:

\vspace{-1em}
\begin{equation}
    \mathcal{L}_{d 2}=\max_{i} \left(\left\|\mathbf{e}_{i}\right\|_{\infty}^{2}-1,0\right)
\end{equation}

The total regularization loss is then:

\vspace{-1em}
\begin{equation}
    \mathcal{L}_{\text{reg}}=\mathcal{L}_{d 1}+\mathcal{L}_{d 2}
    \label{eq:reg_loss_total}
\end{equation}

We define our VQ encoding loss function as the sum of the encoding loss
function from \eqref{eq:vq-vae-loss} and regularization loss from
\eqref{eq:reg_loss_total} scaled by a constant factor,
$\lambda_{\text{reg}}$:

\vspace{-1em}
\begin{equation}
    \begin{split}
        \mathcal{L}_{\text{vq-enc}}=\|s g[E(\mathbf{x})]-\mathbf{e}\|_{2}^{2}+\beta\|s g[\mathbf{e}]-E(\mathbf{x})\|_{2}^{2}\\
        +\lambda_{\text{reg}}\mathcal{L}_{\text{reg}}
    \end{split}
\end{equation}

Our proposed auxiliary task is a typical classification problem, and so we use
Softmax cross entropy loss, given as $\mathcal{L}_\text{class}$, for the
total training loss. Which is given as the following equation: 

\vspace{-1em}
\begin{equation}
    \mathcal{L}_{class}=-\sum_{i}^{N_{a}}a_{i}\log\left(\sigma(C(e))\right)
\end{equation}

\noindent
where $\sigma$ is the softmax function, $a_{i}$ is the action, $N_{a}$ is the number of possible actions, and $C$ is the classification net. 

Thus, the total objective of our VQ-RL training can be
written as:

\vspace{-1em}
\begin{equation}
    \mathcal{L}_{\text{vq-rl}}=\mathcal{L}_{\text{rl}}+\lambda_{\text{vq-enc}}\mathcal{L}_{\text{vq-enc}}+\lambda_{\text{class}}\mathcal{L}_{\text{class}}
\end{equation}

\noindent
where $\mathcal{L}_{\text{rl}}$ is the training loss from the corresponding RL
algorithm, $ \mathcal{L}_{\text{class}} $ is the classification loss in the
auxiliary task, \textit{i.e.}, Softmax cross entropy loss,
$\lambda_{\text{vq-enc}}$ is the weight of the VQ encoding loss, and
$\lambda_{\text{class}}$ is the weight of the classification loss.
$\mathcal{L}_{\text{vq-enc}}$ gathers the features extracted by the feature
extractor around the embeddings in the codebook and ensures a nontrivial
distance between the embeddings. $ \mathcal{L}_{\text{class}} $ leads the training
of VQ encodings to align with policy learning. Therefore, all these endow
VQ-RL with the potential to understand states better and improve the
generalizability and robustness of RL models. We will present our results of testing this
hypothesis in the next section.

    \section{Simulations}
    
This section studies three aspects of VQ-RL, interpretability, robustness,
    and generalization on three domains: CartPole \cite{barto1983neuronlike}, Minigrid \cite{gym_minigrid}, and CoinRun \citep{DBLP:conf/icml/CobbeHHS20}. We
    choose the PPO approach \cite{schulman2017proximal} for our analysis but we
    believe similar conclusions can be easily generalized to other RL methods.
    The hyper-parameter settings and neural networks designs can be found in our appendix.

    \subsection{CartPole}

    The first domain we study is CartPole, in which
    the agent controls a cart which can move to the left or right to maintain
    the balance of a pole stationed on top of it. Our analysis compared
    training three different approaches: PPO using our VQ-RL framework (which
    we call VQ-PPO), VQ-PPO with our two proposed regularization methods (which
    we call VQ-PPO-Reg), and the original PPO algorithm.

\begin{figure}
        \centering
        \includegraphics[width=0.95\columnwidth]{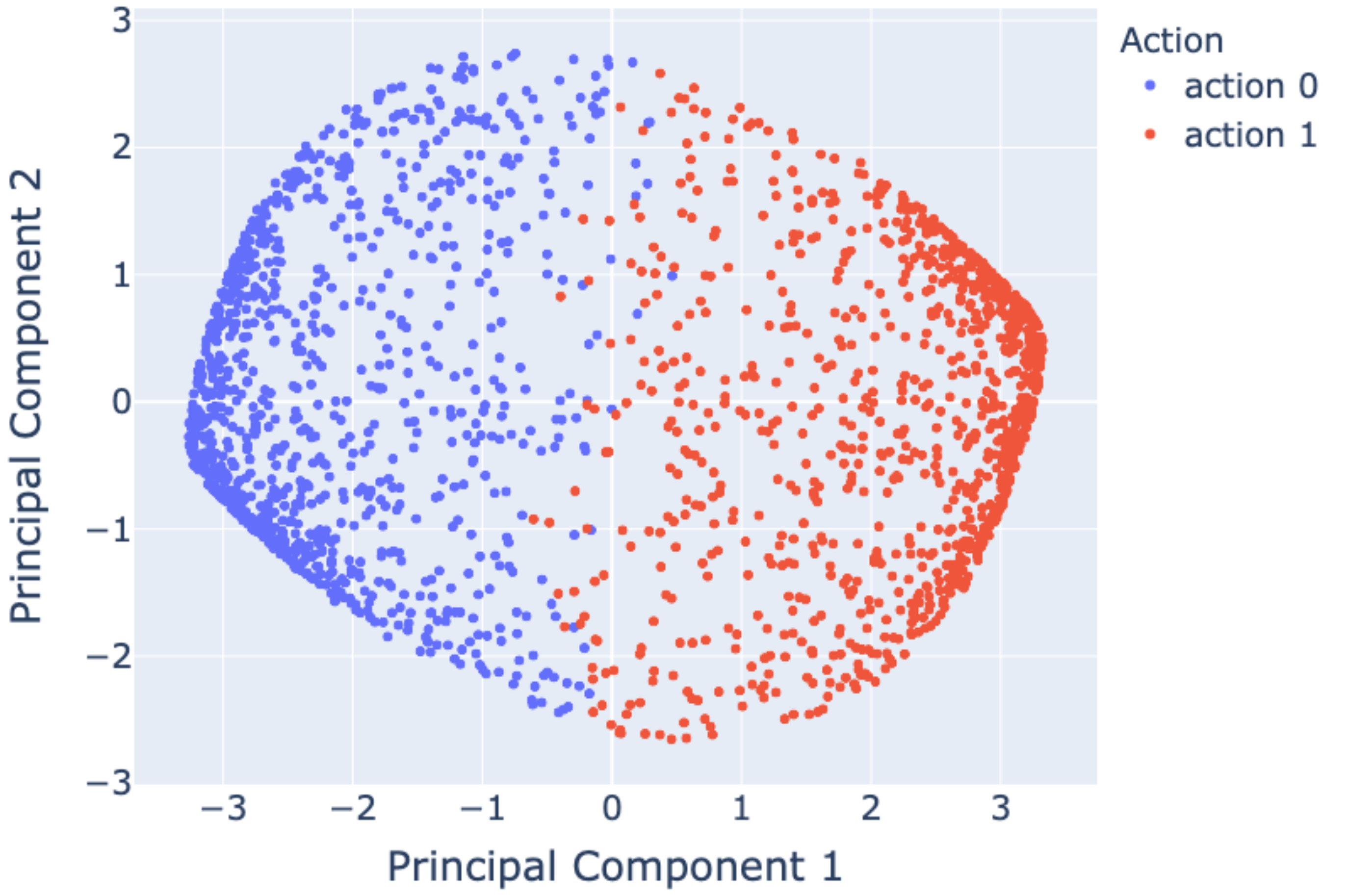}
        \caption{PCA visualization of features learned using the PPO algorithm
        on the CartPole-v1 domain.}
        \label{fig:ppo_cartpole_pca}
    \end{figure}
\begin{figure}
        \centering

        \begin{subfigure}[b]{0.95\columnwidth}
         \centering
         \includegraphics[width=\linewidth]{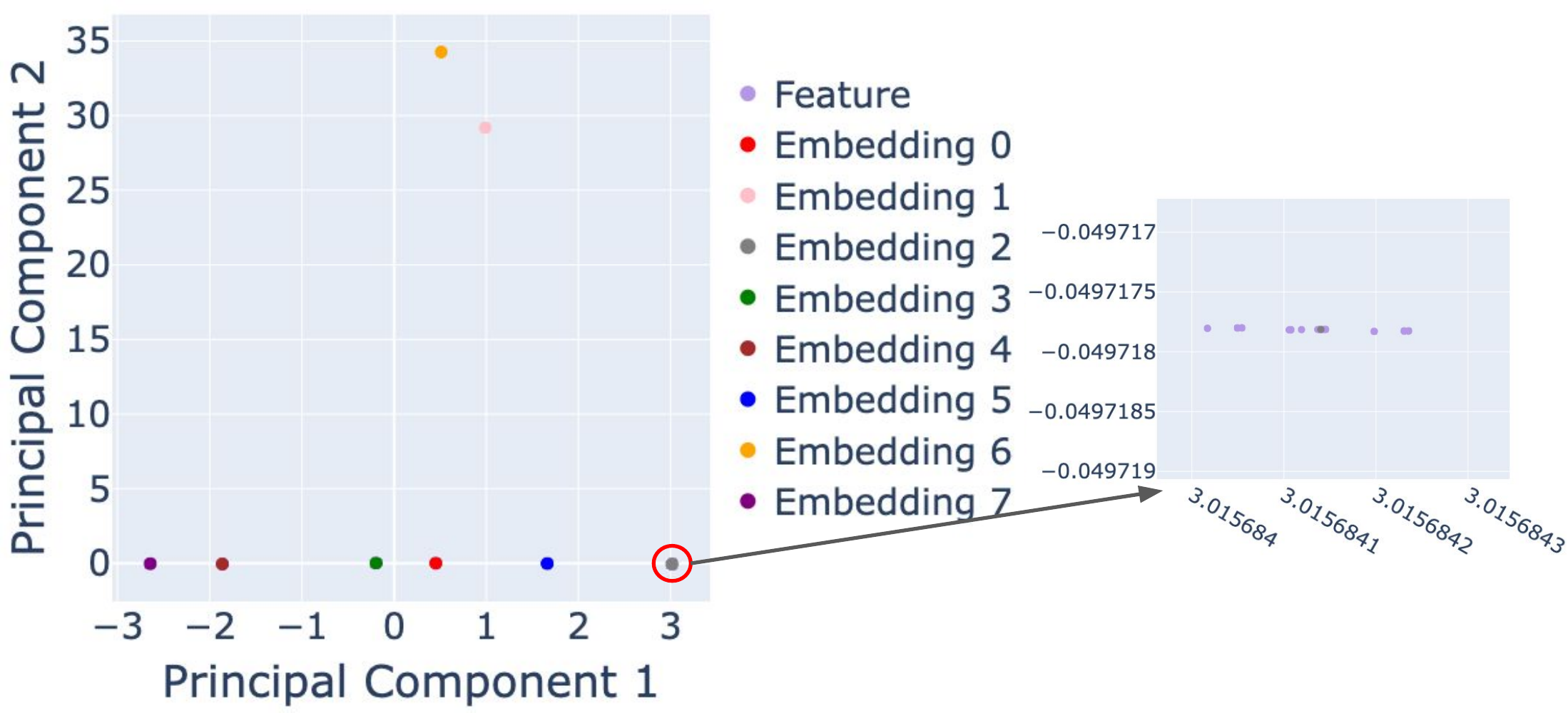}
        \caption{PCA visualization of VQ-PPO}
         \label{fig:vqppo_cartpole_pca}
        \end{subfigure}

        \begin{subfigure}[b]{0.95\columnwidth}
         \centering
         \includegraphics[width=\linewidth]{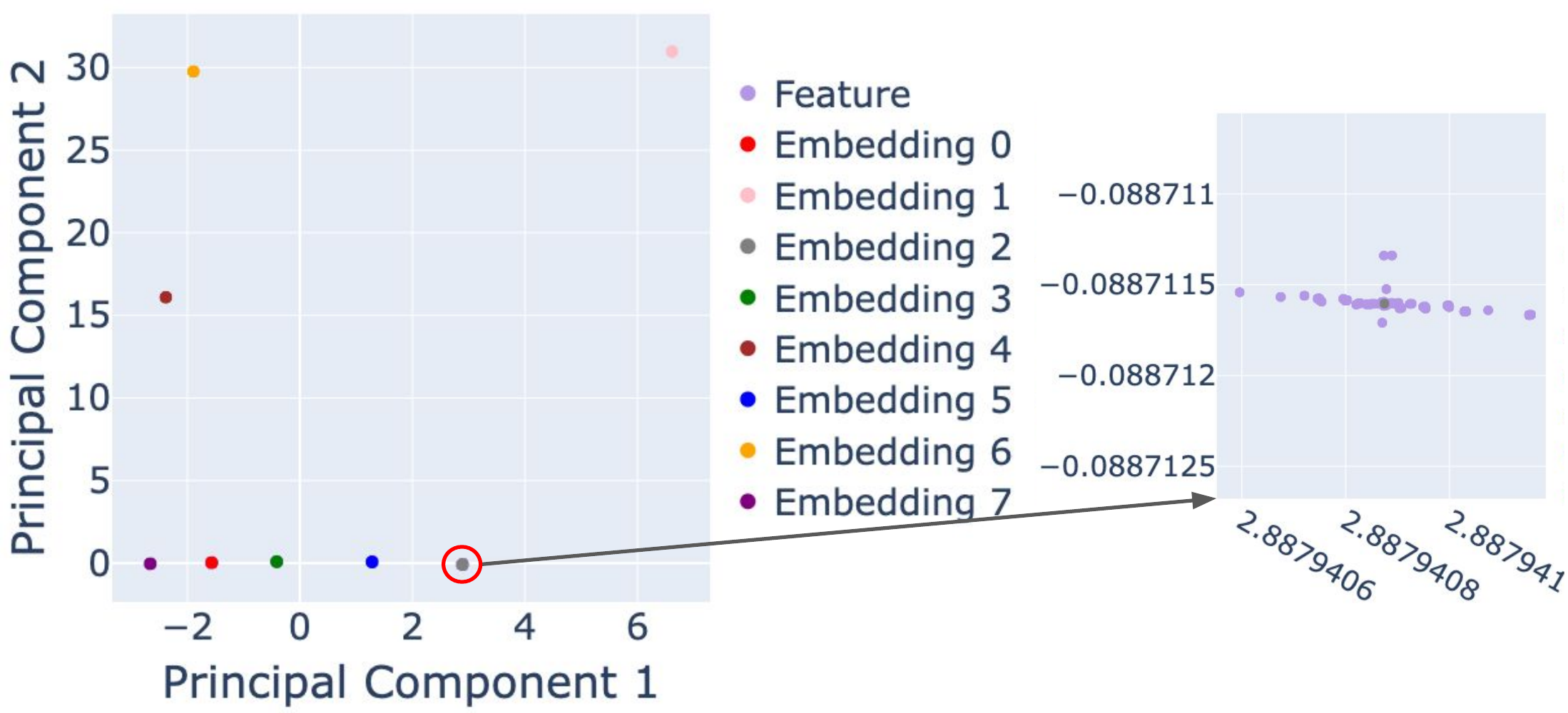}
        \caption{PCA visualization of VQ-PPO-Reg}
         \label{fig:vqppo_cartpole_reg_pca}
        \end{subfigure}

        \caption{%
            Visualizations of features and
            embeddings for the CartPole-v1 domain using VQ-PPO and VQ-PPO-Reg. The right part of each sub-figure zooms in on the point representing
            Embedding 2. The embeddings on the x-axis are tightly distributed with features, while no features surround the other embeddings away from the x-axis.
        }
        \label{fig:vqppo_cartpole}
    \end{figure}

    \paragraph{Interpretability}
    To investigate interpretability we explore the embedding spaces of the
    three contrasting methods. Specifically, we randomly select 2,000 states
    from the legal interval of variables in states of CartPole, and then
    leverage the three feature extractors of trained models to obtain the
    corresponding features. After that, we perform principal component analysis
    (PCA)  on those 2,000 features. The visualization of the PCA results are
    presented in Figure \ref{fig:ppo_cartpole_pca} and
    Figure \ref{fig:vqppo_cartpole}. Since there is no output clustering
    information in the original PPO algorithm, we choose the corresponding
    actions for identification.

    It can be seen that the features obtained by PPO are scattered in a
    relatively large space. Generally speaking, the outputs in the left half
    correspond to action 0, while the outputs in the right half correspond to
    action 1.  However, there is no clear boundary between the two types of
    features in the middle area. As can be seen from Figure \ref{fig:vqppo_cartpole}, although we select
    the number of embeddings in the codebook to be eight, only a subset of the
    embeddings in the codebook are used, all located near the \emph{x}-axis and
    sharing the same principal components. The unused embeddings are highly
    separated from the used embeddings. It is worth noting that since the
    features are tightly clustered around the corresponding embedding, we need
    to zoom in to see the internal distribution. These results are entirely
    consistent with the assumption we made in the introduction, that
    is, using the VQ-RL framework can enhance the clustering properties of
    features.

\begin{figure*}
        \centering
        \includegraphics[width=1.3\columnwidth]{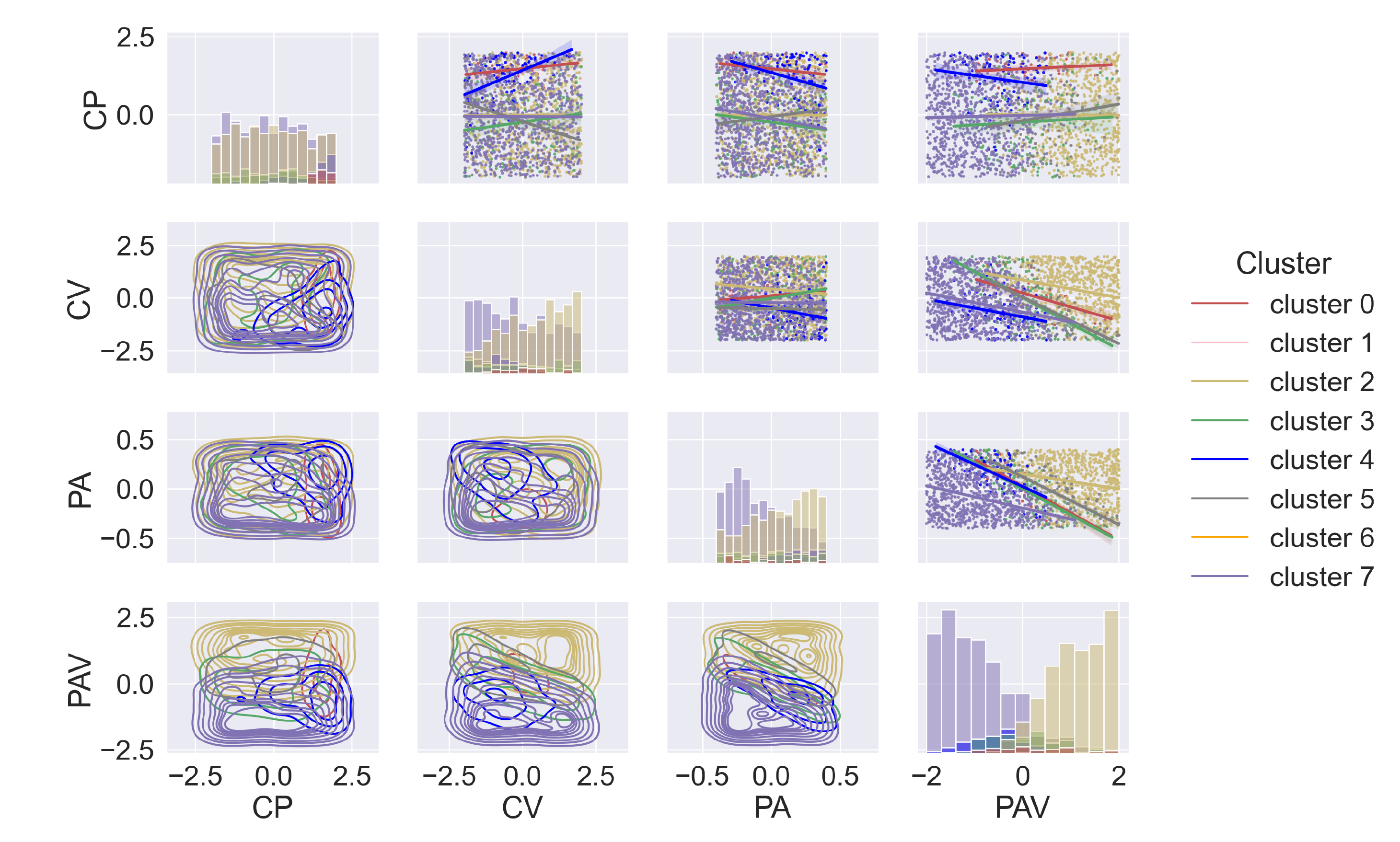}
        \caption{%
            The state clusters for VQ-PPO with our regularization method on the
            CartPole-v1 domain. The abbreviations CP, CV, PA and PAV represent
            cart position, cart velocity, pole angle and pole angular velocity,
            respectively. The upper left plots show the distribution of clusters and the corresponding linear regression lines.  The diagonal presents the histogram graphs. The lower right part is the kernel density estimate (KDE) plots of clusters. We see that PA and PAV are the primary state
            components that distinguish the clusters. In each state component, each cluster presents a continuous distribution in a specific interval.
        }
        \label{fig:vqppo_reg_cartpole_clusters}
    \end{figure*}

    Figure \ref{fig:vqppo_reg_cartpole_clusters} illustrates the clustering of
    the 2,000 states of VQ-PPO-Reg. Since VQ-RL clusters states,  we
    can gain further insight into the internal information of deep models in
    decision-making, such as the priority of state components, the distribution of states in clusters.

    \paragraph{Robustness}

    In the next experiment, we examine the impact of noise on VQ-RL. This is
    common in real-world applications of RL. For example, in the driving of
    unmanned vehicles, rain and snow will produce noise in the visual input.
    Therefore, we need to demonstrate the method’s robustness to noise to
    ensure the safety of RL technology.  Specifically, we add normally
    distributed noise $n\sim\mathcal{N}(0,\, 1)/\delta$ to
    the input states of CartPole, where $\delta$ is a scaling factor to control
    the level of noise.

    \begin{table}
        \centering\small\scshape
        \begin{tabular}{cccc}
            \toprule
            $\delta$ & PPO & VQ-PPO & VQ-PPO-Reg \\
            \midrule
            120 & \textbf{500.0$\pm$0.0} & \textbf{500.0$\pm$0.0} & \textbf{500.0$\pm$0.0}    \\
            110 & 499.4$\pm$2.4          & \textbf{500.0$\pm$0.0} & \textbf{500.0$\pm$0.0}   \\
            100 & 499.5$\pm$2.1          & \textbf{500.0$\pm$0.0} & \textbf{500.0$\pm$0.0}   \\
            90  & 499.3$\pm$2.8          & \textbf{500.0$\pm$0.0} & \textbf{500.0$\pm$0.0}             \\
            80  & 490.7$\pm$28.5         & \textbf{500.0$\pm$0.0} & \textbf{500.0$\pm$0.0}            \\
            70  & 455.8$\pm$75.4         & 495.6$\pm$19.1         & \textbf{500.0$\pm$0.0}            \\
            60  & 394.6$\pm$121.0        & 493.4$\pm$21.4         & \textbf{494.1$\pm$25.9}          \\
            50  & 283.3$\pm$159.6        & 343.3$\pm$167.5        & \textbf{439.9$\pm$128.2}         \\
            \bottomrule
        \end{tabular}
        \caption{%
            The results for the noisy CartPole-v1 test.}
        \label{tab:cartpole_robustness_test}
    \end{table}

    We test the robustness of the three methods on 20 episodes with randomly
  initial states. The results for different levels of noise are
    shown in Table \ref{tab:cartpole_robustness_test}. The introduction of VQ
    encoding and our regularization methods significantly improves the
    robustness of PPO, with the regularized VQ-PPO performing better than
    VQ-PPO without regularization.

    \paragraph{Generalization} To study the generalizability of VQ-RL, we
    propose a modified version of CartPole and name it Gen-CartPole. In this
    environment, we consider changing three game-related parameters that affect
    the game’s transition function between states: the mass of the cart
    ($m_c$), the mass of the pole ($m_p$), and the length of the pole ($l$).
    In each episode in training, the parameter values will be randomly selected
    from among the three training sets shown in the upper table of
    Table \ref{tab:cartpole_gen_test}. It is worth noting that we take a window of the last four consecutive
    states as input states for models, in order recognize the current dynamics.
    After completing training, we test the performance of the three models on
    the test sets in the bottom table of Table \ref{tab:cartpole_gen_test} which also shows their results.
    We see that VQ-PPO-Reg performs better than the other two
    algorithms.



    \begin{table}
        \setlength\tabcolsep{1.3pt}
        \centering\small\scshape
        \begin{tabular}{lccc}
        \toprule
        \multicolumn{4}{c}{Training Sets $\left(m_c, m_p, l\right)$:} \\
            \midrule
            \multicolumn{4}{c}{(0.5, 0.05, 0.25) \ , \ (1.0, 0.1, 0.5) \ , \ (2, 0.2, 1)} \\
            \bottomrule
            \toprule
            \multicolumn{4}{c}{Test Set Results} \\
            \midrule
            $\left(m_c, m_p, l\right)$& PPO & VQ-PPO & VQ-PPO-Reg  \\
            \midrule
            (0.75, 0.075, 0.375) & \textbf{500.0$\pm$0.0} & 495.8$\pm$0.0          & \textbf{500.0$\pm$0.0}    \\
            (1.5, 0.15, 0.75)    & 499.4$\pm$2.4          & \textbf{500.0$\pm$0.0} & \textbf{500.0$\pm$0.0}             \\
            (3, 0.3, 1.5)        & 78.1$\pm$21.5         & 474.4$\pm$29.8         & \textbf{500.0$\pm$0.0}             \\
            (5.0, 0.5, 2.5)      & 45.9$\pm$6.5          & 231.6$\pm$46.2         & \textbf{500.0$\pm$0.0}             \\
            (7.5, 0.75, 3.75)    & 65.7$\pm$11.7         & 190.9$\pm$49.4         & \textbf{363.2$\pm$168.7}           \\
            \bottomrule
        \end{tabular}
        \caption{This table shows the training set parameters and the results for the Gen-CartPole-v1 test sets along with their parameter sets.}
        \label{tab:cartpole_gen_test}
    \end{table}

    \subsection{MiniGrid-Empty}

    The MiniGrid-Empty-Random-6x6-v0 domain is chosen for the
    model training in the interpretability and robustness experiments in this
    section. In this domain, an agent moves from a random initial position and
    orientation to a green target in the lower right corner, using a set of
    three actions: turning left, turning right, and moving forward. The state
    of the agent is encoded using a 7$\times$7 field of view (FoV) with three
    layers of channels, where the first layer encodes object identifiers
    ranging from 0 to 10, the second layer encodes object colors as integers
    from 0 to 5, and the third layer encodes whether the block is empty or not
    (0 or 1). Each result in this section is calculated using 50 episodes.

\begin{figure}
        \centering
        \includegraphics[scale=0.32]{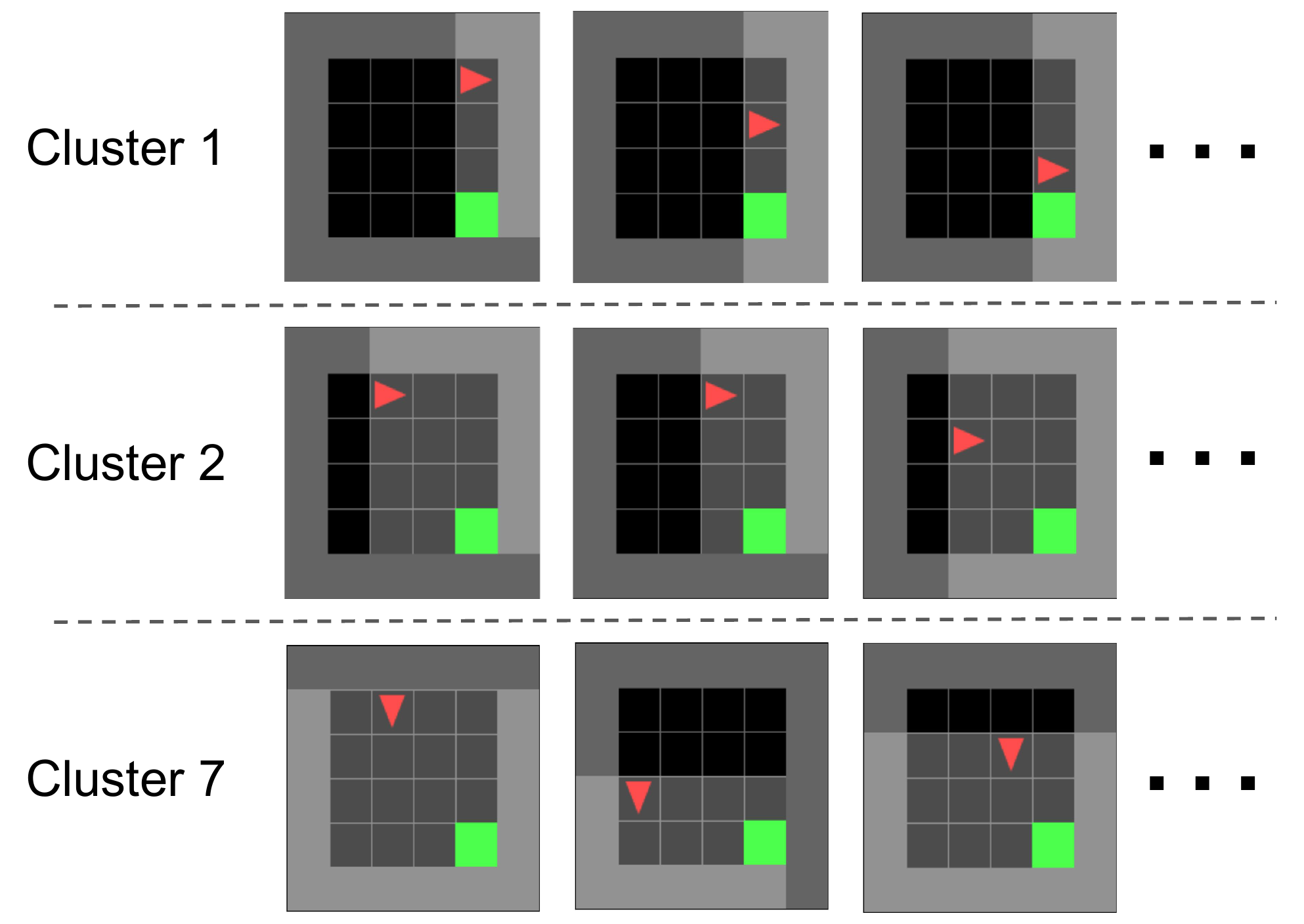}
        \caption{%
            The clustering result of VQ-PPO-Reg on
            MiniGrid-Empty-Random-6x6-v0. The model divides the state into
            three clusters according to the actions that need to be performed.
            In cluster 1, the agent performs the
            turning right action. Cluster 2 represents moving forward. Cluster 7 is turning left.
        }
        \label{fig:minigrid_clusters}
    \end{figure}

    \paragraph{Interpretability} Figure \ref{fig:minigrid_clusters} shows the
    clustering results of the state of the trained VQ-PPO-Reg model. The model
    categorizes the state space into three corresponding clusters according to
    the three actions performed in the policy.

    \paragraph{Robustness}

    The robustness of the model is tested by adding random noise to the state
    (the FoV component) in MiniGrid-Empty-Random-6x6-v0. Specifically, as
    performing each action, each grid cell in the FoV has a certain probability
    $p_a$ of being `attacked', which results in an ``illusion", that is, each
    of the three encoding layers assigns a random value within the valid
    encoding range for this grid cell.  The results are given in
    Table \ref{tab:minigrid_noisy_test}. When $p_a$ is small, the performance of
    the three models is close. However, as $p_a$ increases, the performance of
    VQ-PPO decreases rapidly, but VQ-PPO-Reg can still maintain a relative
    advantage over the other two models.

    \begin{table}
        \centering\small\scshape
        \begin{tabular}{cccc}
            \toprule
            $p_a$ & PPO & VQ-PPO &  VQ-PPO-Reg  \\
            \midrule
            0.01 & 0.96$\pm$0.01          & \textbf{0.97$\pm$0.01} & 0.96$\pm$0.02          \\
            0.05 & 0.94$\pm$0.04          & \textbf{0.95$\pm$0.04}          & \textbf{0.95$\pm$0.04} \\
            0.1  & 0.89$\pm$0.08          & \textbf{0.93$\pm$0.06} & 0.91$\pm$0.08          \\
            0.2  & 0.82$\pm$0.18          & 0.83$\pm$0.20 & \textbf{0.83$\pm$0.18}          \\
            0.3  & 0.76$\pm$0.25.         & 0.66$\pm$0.34          & \textbf{0.78$\pm$0.22} \\
            0.4  & \textbf{0.74$\pm$0.27} & 0.44$\pm$0.42          & 0.72$\pm$0.28           \\
            0.5  & 0.60$\pm$0.38          & 0.38$\pm$0.42          & \textbf{0.67$\pm$0.34} \\
            \bottomrule
        \end{tabular}
        \caption{The results for the noisy Minigrid test.}
        \label{tab:minigrid_noisy_test}
    \end{table}

\begin{figure}
        \centering
        \includegraphics[width=0.8\columnwidth]{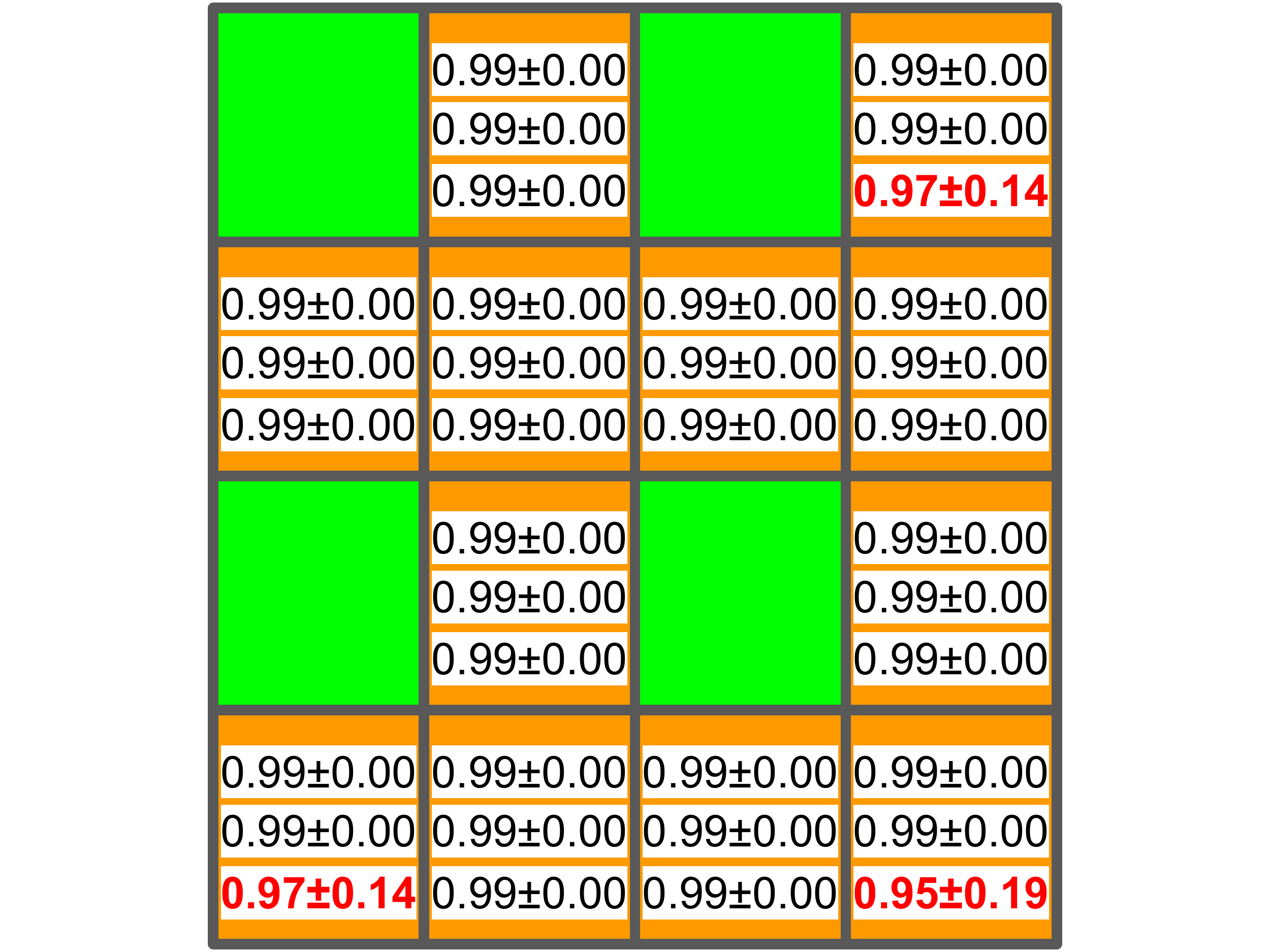}
        \caption{Grid showing the locations of goals in training and test and
        the generalization test results. In the label for each orange cell, the first,
        second, and third row correspond to the performance of the VQ-PPO, VQ-PPO-Reg,
        and PPO methods respectively.}
        \label{fig:gen_minigrid}
    \end{figure}

    \paragraph{Generalization} To test the generalization of our models, we
    propose a modified MiniGrid-Empty-Random-6x6-v0 domain that we call
    \emph{Gen-MiniGrid-Empty-Random-v0}. During training, each episode randomly produces
    the goal from four possible locations (the green cells in
    Figure \ref{fig:gen_minigrid}). In the evaluation, we test the performance of
    models in the rest of the locations (the orange cells in
    Figure \ref{fig:gen_minigrid}). The corresponding results are also presented in
    Figure \ref{fig:gen_minigrid}. We can see that the three methods have very
    similar performance, except for the three corner cells, for which the PPO
    algorithm performs decidedly worse than the other two.

    \subsection{CoinRun}

    In this experiment, we leverage the Procgen version of CoinRun
    \citep{DBLP:conf/icml/CobbeHHS20} for testing. In this game, the agent needs
    to move to the far-right of the screen and touch the gold coin to get
    rewards. We used 200 generated levels of CoinRun on the easy difficulty setting to train our model
    and the same network and hyperparameters as the one in the original paper \citep{DBLP:conf/icml/CobbeHHS20}, except for adding VQ
    encoding. Furthermore, each model is trained three times with different
    random seeds, and the evaluation of each model is based on the average of
    the three instances.

\begin{figure}
        \centering
        \includegraphics[width=\linewidth]{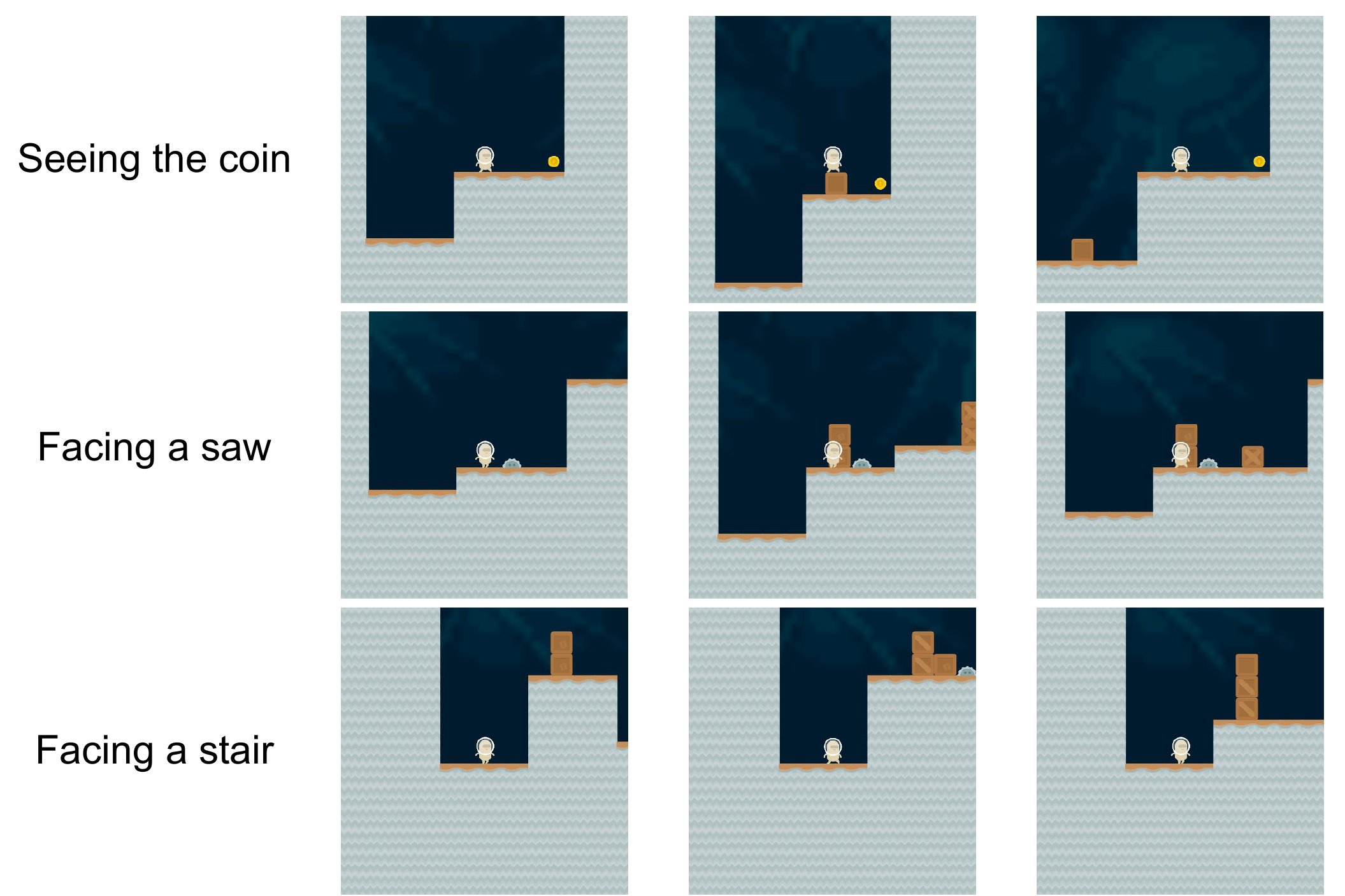}
        \caption{Some sub-classified state examples of VQ-PPO-Reg on CoinRun.}
        \label{fig:coinrun_clusters}
    \end{figure}

    \paragraph{Interpretability} The same cluster after training
    may contain multiple different highest probability actions, although the
    classification loss function constrains features in the same cluster to
    have the same highest probability actions as much as possible. This may be
    due to the choice of hyperparameters (such as relatively small weight for
    the loss of the classification task, etc.) and the action distribution of
    states with high entropy (some states may have multiple optimum actions to
    achieve the goal.). However, we can see meaningful clustering results if
    the states are sub-classified according to different highest probability
    actions within each cluster. For example, as shown in
    Figure \ref{fig:coinrun_clusters}, the states in each row originate from the same
    cluster and have the same highest probability action. Therefore, the
    clustering in VQ-RL can still explore the internal state of the models and
    help us further understand the decision-making process in deep RL.

    \paragraph{Robustness}
    \begin{table}
        \centering\small\scshape
        \tabcolsep=0.1cm
        \begin{tabular}{cccc}
            \toprule
            $p$ & PPO & VQ-PPO & VQ-PPO-Reg  \\
            \midrule
            1  &  9.43$\pm$2.31 &  9.50$\pm$2.17(+0.7\%) &  \textbf{9.63$\pm$1.87(+2.1\%)}  \\
            10 &  8.60$\pm$3.46 &  \textbf{8.93$\pm$3.08(+3.8\%)} &  8.90$\pm$3.12(+3.1\%)  \\
            20 &  7.36$\pm$4.40 &  7.66$\pm$4.22(+4.1\%) &  \textbf{8.63$\pm$3.43(+17\%)}  \\
            30 &  6.53$\pm$4.75 &  \textbf{7.53$\pm$4.31(+15\%)} &  7.33$\pm$4.42(+12\%)  \\
            40 &  6.40$\pm$4.80 &  6.56$\pm$4.74(+2.5\%) &  \textbf{6.70$\pm$4.70(+4.7\%)}  \\
            50 &  6.03$\pm$4.89 &  5.80$\pm$4.93(-3.9\%) &  \textbf{6.33$\pm$4.81(+5.0\%)}  \\
            \bottomrule
        \end{tabular}
        \caption{The results for the noisy CoinRun test. The proportion of
        noisy points is given by $p\times 0.004$, where $p$ is a configurable
        parameter. We also include percent improvements from the baseline.}
        \label{tab:coinrun_noisy_test}
    \end{table}

    Salt and Pepper (SP) is one of the most common image noises
    \cite{DBLP:journals/tip/ChanHN05}. Therefore, we inject SP noise with
    different degrees to the RGB states in CoinRun to test the robustness of
    the three models. The results are shown in
    Table \ref{tab:coinrun_noisy_test}. We see that the VQ-RL framework
     improves the robustness of PPO.

    \paragraph{Generalization}
     \begin{table}[H]
        \centering\small\scshape
        \tabcolsep=0.1cm
        \begin{tabular}{cccc}
            \toprule
            Run    & PPO & VQ-PPO & VQ-PPO-Reg  \\
            \midrule
            1   &  8.77$\pm$3.28 &  8.58$\pm$3.49(-3.2\%) &  \textbf{8.84$\pm$3.20(+0.8\%)}  \\
            2   &  8.59$\pm$3.48 &  8.53$\pm$3.54(-0.7\%) &  \textbf{8.62$\pm$3.44(+0.3\%)}  \\
            3   &  8.77$\pm$3.28 &  8.66$\pm$3.40(-1.3\%) &  \textbf{9.00$\pm$3.00(+2.6\%)}  \\
            \midrule
            Ave. &  8.71$\pm$3.35 & 8.59$\pm$3.48(-1.4\%) & \textbf{8.82$\pm$3.22(+1.2\%)}  \\
            \bottomrule
        \end{tabular}
        \caption{The results for the generalization CoinRun test, including percent changes from the baseline PPO.}
        \label{tab:coinrun_gen_test}
    \end{table}

    In the final experiments, we test the generalization of the models for the
    CoinRun domain. We test the three models trained with the same random
    parameters on the same 1,000 unseen `all distributions' episodes to ensure
    fairness. We performed this experiment three times, labeled as Run1-3. The results for this generalization experiment can be seen in Table \ref{tab:coinrun_gen_test}. We see that the VQ-RL framework slightly improves the generalization of PPO.


    \section{Discussion and Conclusion}
    
    In this paper, we propose a new framework to enhance the clustering
    properties of the embedding space of deep reinforcement learning methods.
    In the experimental part, we test the effectiveness of
    VQ-RL. Our results on three test domains suggest
    that 1) VQ-RL improves the interpretability of deep RL from different aspects in the three domains; 2) regarding robustness and generalization, VQ-RL has significant advantages in CartPole with the continuous representation and different degrees of improvement in the other two domains using hand-coding and visual inputs. In general, the baseline is improved in these two aspects; 3) the regularization methods are effective and essential to maintain the robustness and generalization in VQ-RL.


    \section{Acknowledgments}
    Research was sponsored by the Army Research Office and was accomplished under Grant Number W911NF-20-1-0002. The views and conclusions contained in this document are those of the authors and should not be interpreted as representing the official policies, either expressed or implied, of the Army Research Office or the U.S. Government. The U.S. Government is authorized to reproduce and distribute reprints for Government purposes notwithstanding any copyright notation herein.

\bibliography{aaai23}

\end{document}